\documentclass[11pt,a4paper]{article}
\usepackage[hyperref]{acl2018}
\usepackage{times}
\usepackage{latexsym}
\usepackage{footnote}
\usepackage{tablefootnote}
\usepackage{enumitem}

\usepackage{bm}
\usepackage{multirow}
\usepackage{amsfonts}
\usepackage[boxed,linesnumbered,lined]{algorithm2e}
\usepackage{algpseudocode}
\usepackage{graphicx}
\usepackage{hyphenat}
\usepackage{verbatim}

\usepackage{CJKutf8}
\usepackage{graphicx}
\usepackage{nonfloat}
\usepackage{fmtcount}
\usepackage{threeparttable}
\usepackage{caption}

\usepackage{url}

\aclfinalcopy 

\title{DuReader: a Chinese Machine Reading Comprehension Dataset from Real-world Applications}

\author{ \textbf{Wei He, Kai Liu, Jing Liu, Yajuan Lyu, Shiqi Zhao, Xinyan Xiao, Yuan Liu, Yizhong Wang,} \\
			\textbf{Hua Wu, Qiaoqiao She, Xuan Liu, Tian Wu, Haifeng Wang}\\
	Baidu Inc., Beijing, China \\
	\{hewei06, liukai20, liujing46, lvyajuan, zhaoshiqi, xiaoxinyan, liuyuan04, wangyizhong01, \\
	wu\_hua, sheqiaoqiao, liuxuan, wutian, wanghaifeng\}@baidu.com }

\date{}

\begin{document}
\begin{CJK*}{UTF8}{gbsn}

\maketitle
\begin{abstract}
This paper introduces DuReader, a new large-scale, open-domain Chinese machine reading comprehension (MRC) dataset, designed to address real-world MRC. DuReader has three advantages over previous MRC datasets: (1) {\bf data sources:} questions and documents are based on Baidu Search and Baidu Zhidao\footnote{Zhidao (\url{https://zhidao.baidu.com}) is the largest Chinese community-based question answering (CQA) site in the world.}; answers are manually generated. (2) {\bf question types:} it provides rich annotations for more question types, especially yes-no and opinion questions, that leaves more opportunity for the research community. (3) {\bf scale:} it contains 200K questions, 420K answers and 1M documents; it is the largest Chinese MRC dataset so far. Experiments show that human performance is well above current state-of-the-art baseline systems, leaving plenty of room for the community to make improvements.  To help the community make these improvements, both DuReader\footnote{\url{http://ai.baidu.com/broad/download?dataset=dureader}} and baseline systems\footnote{\url{https://github.com/baidu/DuReader}} have been posted online. We also organize a shared competition to encourage the exploration of more models. Since the release of the task, there are significant improvements over the baselines. 

\end{abstract}
\section{Introduction} \label{section:intro}

The task of machine reading comprehension (MRC) aims to empower machines to answer questions after reading articles~\cite{rajpurkar2016squad,nguyen2016marco}. In recent years, a number of datasets have been developed for MRC, as shown in Table \ref{table:vs_dureader}.  These datasets have led to advances such as Match-LSTM~\cite{wang2017machine}, BiDAF~\cite{seo2016bidirectional}, AoA Reader~\cite{cui2016attention}, DCN~\cite{xiong2016dynamic} and R-Net~\cite{wang2017gated}. This paper hopes to advance MRC even further with the release of  DuReader, challenging the community to deal with more realistic data sources, more types of questions and more scale, as illustrated in Tables 1-4. Table~\ref{table:vs_dureader} highlights DuReader's advantages over previous datasets in terms of data sources and scale. Tables 2-4 highlight DuReader's advantages in the range of questions.

\begin{table*}[!t]\footnotesize
\captionsetup{font=footnotesize}
\centering
\begin{threeparttable}
\begin{tabular}{ccccccc}
Dataset & Lang & \#Que. & \#Docs & Source of Que. & Source of Docs & Answer Type\\
\hline
CNN/DM~\cite{hermann2015teaching}& EN & 1.4M & 300K & Synthetic cloze & News & Fill in entity \\
HLF-RC~\cite{Cui2016Consensus}& ZH & 100K & 28K & Synthetic cloze & Fairy/News & Fill in word \\
CBT~\cite{hill2015goldilocks}  & EN & 688K & 108 & Synthetic cloze & Children's books & Multi. choices \\
RACE~\cite{lai2017large}  & EN & 870K & 50K & English exam & English exam & Multi. choices \\ 
MCTest~\cite{richardson2013mctest}  & EN & 2K & 500 & Crowdsourced & Fictional stories & Multi. choices \\
NewsQA~\cite{trischler2017:newsqa}  & EN & 100K & 10K & Crowdsourced & CNN & Span of words \\
SQuAD~\cite{rajpurkar2016squad}  & EN & 100K & 536 & Crowdsourced & Wiki. & Span of words \\
SearchQA~\cite{dunn2017searchqa}  & EN & 140K & 6.9M & QA site & Web doc. & Span of words \\
TrivaQA~\cite{joshi2017triviaqa}  & EN & 40K & 660K & Trivia websites & Wiki./Web doc. & Span/substring of words \\
NarrativeQA~\cite{kovcisky2017narrativeqa} & EN & 46K & 1.5K & Crowdsourced & Book\&movie & Manual summary \\
MS-MARCO~\cite{nguyen2016marco}  & EN & 100K & 200K\tnote{1} & User logs & Web doc. & Manual summary \\
\hline
\textbf{DuReader (this paper)} & \textbf{ZH} & \textbf{200k} & \textbf{1M} & \textbf{User logs} & \textbf{Web doc./CQA} &\textbf{Manual summary} \\
\hline
\end{tabular}
\caption{DuReader has three advantages over previous MRC datasets: (1)
{\bf data sources}: questions and documents are based on Baidu Search \& Baidu Zhidao; answers are manually generated, (2) {\bf question types}, and (3) {\bf scale}:
200k questions, 420k answers and 1M documents (largest Chinese MRC dataset so far).  The next three tables address advantage (2).} \label{table:vs_dureader}
\begin{tablenotes}
\item[1] Number of unique documents
\end{tablenotes}
\end{threeparttable}
\end{table*} 

Ideally, a good dataset should be based on questions from real applications. However, many existing datasets have been forced to make various compromises such as: (1) {\bf cloze task:}
Data is synthesized missing a keyword.  The task is to fill in the missing keyword \cite{hermann2015teaching,Cui2016Consensus,hill2015goldilocks}. (2) {\bf multiple-choice exams:} \newcite{richardson2013mctest} collect both fictional stories and the corresponding multiple-choice questions by crowdsourcing. \newcite{lai2017large} collect the multiple-choice questions from English exams. (3) {\bf crowdsourcing:} Turkers are given documents (e.g., articles from the news and/or Wikipedia) and are asked to construct questions after reading the documents\cite{trischler2017:newsqa,rajpurkar2016squad,kovcisky2017narrativeqa}. 


The limitations of the datasets lead to build datasets based on queries that real users submitted to real search engines. MS-MARCO \cite{nguyen2016marco} is based on Bing logs (in English), and DuReader (this paper) is based on the logs of Baidu Search (in Chinese). Besides {\bf question sources}, DuReader complements MS-MARCO and other datasets in the following ways:

{\bf question types:}
DuReader contains a richer inventory of questions than previous datasets. Each question was manually annotated as either Entity, Description or YesNo and one of Fact or Opinion. In particular, it annotates yes-no and opinion questions that take a large proportion in real user’s questions. Prior work has largely emphasized facts, but DuReader are full of opinions as well as facts. Much of the work on question answering involves span selection, methods that answer questions by returning a single substring extracted from a single document. Span selection may work well for factoids (entities), but it is less appropriate for yes-no questions and opinion questions (especially when the answer involves a summary computed over several different documents).

{\bf document sources:}
DuReader collects documents from the search results of Baidu Search as well as Baidu Zhidao. All the content in Baidu Zhidao is generated by users, making it different from the common web pages. It is interesting to see if solutions designed for one scenario (search) transfer easily to another scenario (question answering community). Additionally, previous work provides only a single paragraph~\cite{rajpurkar2016squad} or a few passages~\cite{nguyen2016marco} to extract or generate answers, while DuReader provides multiple full documents (that contains a lot of paragraphs or passages) for each question to generate answers. This will raise paragraph selection (i.e. select the paragraphs likely containing answers) an important challenge as shown in Section 4. 

{\bf data scale:}
The first release of DuReader contains 200K questions, 1M documents and more than 420K human-summarized answers. To the best of our knowledge, DuReader is the largest Chinese MRC dataset so far. 


\begin{table*}[!t]\centering
\begin{tabular}{l|ll}
\hline
{}  & 	\textbf{Fact} & \textbf{Opinion} \\ 
\hline
\textbf{Entity} & iphone哪天发布 & 2017最好看的十部电影 \\
{}				& On which day will iphone be released & Top 10 movies of 2017 \\
\hline
\textbf{Description} & 消防车为什么是红的 & 丰田卡罗拉怎么样   \\ 
{}				& Why are firetrucks red & How is Toyota Carola \\
\hline
\textbf{YesNo} 	& 39.5度算高烧吗  & 学围棋能开发智力吗 \\
{}				& Is 39.5 degree a high fever & Does learning to play go improve intelligence \\
\hline
\end{tabular}
\caption{Examples of the six types of questions in Chinese (with glosses in English). Previous datasets have focused on fact-entity and fact-description, though all six types are common in search logs.}
\label{table:quest_types_example}
\end{table*}

\begin{table}[!t]
\centering
\begin{tabular}{lccc}
\hline
\textbf{}    			& \textbf{Fact} & \textbf{Opinion} 	& \textbf{Total} \\ 
\hline
\textbf{Entity} 		& 23.4\%		& 8.5\% 			& 31.9\%	\\ 
\textbf{Description} 	& 34.6\% 		& 17.8\%  			& 52.5\%	\\ 
\textbf{YesNo} 		& 8.2\% 		& 7.5\%   			& 15.6\%	\\
\hline
\textbf{Total} 		& 66.2\% 		& 33.8\%   			& 100.0\%	\\
\hline
\end{tabular}
\caption{Pilot Study found that all six types of question queries are common in search logs. Previous MRC datasets have emphasized span-selection methods. Such methods are appropriate for fact-entity and fact-description. Opinions and yes-no leave big opportunities (about 33.8\% and 15.6\% of the sample, respectively).}
\label{table:dis_quest_type_baidu}
\end{table}

\section{Pilot Study} \label{section:question_type}

What types of question queries do we find in the logs of a search engine?  A pilot study was performed to create a taxonomy of question types. We started with a relatively small sample of 1000 question queries, selected from a single day of Baidu Search logs.

The pilot helped us to agree on the following taxonomy of question types.  
Each question was manually annotated as:
\begin{itemize}[topsep=0pt,itemsep=-1ex,partopsep=1ex,parsep=1ex]
\item
either \emph{Fact} or \emph{Opinion}, and
\item
one of: \emph{Entity}, \emph{Description} or \emph{YesNo}
\end{itemize}

Regarding to \emph{Entity} questions, the answers are expected to be a single entity or a list of entities. While the answers to \emph{Description} questions are usually multi-sentence summaries. The \emph{Description} questions contain how/why questions, comparative questions that comparing two or more objects, and the questions that inquiring the merits/demerits of goods, etc. As for \emph{YesNo} questions, the answers are expected to be an affirmative or negative answers with supporting evidences. After the deep analysis of the sampled questions, we find that whichever the expected answer type is, a question can be further classified into \emph{Fact} or \emph{Opinion}, depending on whether it is about asking a fact or an opinion. Table \ref{table:quest_types_example} gives the examples of the six types of questions. 

The pilot study helped us identify a number of important issues. Table \ref{table:dis_quest_type_baidu} shows that all six types of question queries are common in the logs of Baidu Search, while previous work has tended to focus on fact-entity and fact-description questions. As shown in Table \ref{table:dis_quest_type_baidu}, fact-entity questions account for a relatively small fraction (23.4\%) of the sample. Fact-descriptions account for a larger fraction of the sample (34.6\%). From this Table, we can see that opinions (33.8\%) are common in search logs. Yes-No questions account for 15.6\%, with one half about fact, another half about opinion. 

Previous MRC datasets have emphasized span-selection methods. Such methods are appropriate for fact-entity and fact-description, but it is problematic when the answer involves a summary of multiple sentences from multiple documents, especially for Yes-no and opinion questions. This requires methods that go beyond currently popular methods such as span selection, and leave large opportunity for the community. 


\section{Scaling up from the Pilot to DuReader}
\label{section:dureader_dataset}
\subsection{Data Collection and Annotation}

\subsubsection{Data Collection}
After the successful completion of the pilot study, we began work on scaling up the relatively small sample of 1k questions to a more ambitious collection of 200k questions.  

The DuReader is a sequence of 4-tuples: 
\{\emph{q}, \emph{t}, \emph{D}, \emph{A}\}, 
where \emph{q} is a question, \emph{t} is a question type,
\emph{D} is a set of relevant documents, and \emph{A} is an answer set produced by human annotators.

Before labeling question types, we need to collect a set of questions $q$ from search logs. According to our estimation, there are about $21\%$ question queries in search logs. It would take too much time, if human annotators manually label each query in search logs. Hence, we first randomly sample the most frequent queries from search logs, and use a pre-trained classifier (with recall higher than $90\%$) to automatically select question queries from search logs. Then, workers will annotate the question queries selected by the classifier. Since this annotation task is relatively easy, each query was annotated by one worker. The experts will further review all the annotations by workers and correct them if the annotation is wrong. The accuracy of workers' annotation (judged by experts) is higher than $98\%$. 


Initially, we have 1M frequent queries sampled from search logs. The classifier automatically selected 280K question queries. After human annotation, there are 210K question queries left. Eventually, we uniformly sampled 200K questions from the 210K question queries. 


We then collect the relevant documents, $D$, by submitting questions to two sources, Baidu Search and Baidu Zhidao. Note that the two sources are very different from one another; Zhidao contains user-generated content and tends to have more documents relevant to opinions. Since the two sources are so different from each another, we decided to randomly split the 200k unique questions into two subsets.  The first subset was used to produce the top 5 ranked documents from one source, and the second subset was used to produce the top 5 ranked documents from the other source. 

We also believe that it is important to keep the entire document unlike previous work which kept a single paragraph~\cite{rajpurkar2016squad} or a few passages~\cite{nguyen2016marco}. In this case, paragraph selection (i.e. select the paragraphs likely containing answers) becomes critical to the MRC systems as we will show in Section 4. 

Documents are parsed into a few fields including title and main content. Text has been tokenized into words using a standard API.\footnote{\url{http://ai.baidu.com/tech/nlp/lexical}}

\subsubsection{Question Type Annotation} \label{section:qust_type_anot}

As mentioned above, annotators labeled each question in two passes.  The first pass classified questions into one of three types: \emph{Entity}, \emph{Description} and \emph{YesNo} questions.  The second pass classified questions as either \emph{Fact} or \emph{Opinion}.

Statistics on these classifications are reported in Table 4. Note that these statistics are similar to those reported for the pilot study (Table 3), but different because duplicates were removed from Table 4 (but not from Table 3). We don't want to burden the annotators with lots of copies of the most frequent questions, hence we kept unique questions in DuReader. That said, both tables agree on a number of important points. As pointed out above, previous work has tended to focus on fact-entity and fact-description, while leaves large opportunity on yes-no and opinion questions.

\begin{table}[]
\centering
\begin{tabular}{lccc}
\hline
\textbf{}    			& \textbf{Fact} & \textbf{Opinion} 	& \textbf{Total} \\ 
\hline
\textbf{Entity} 		& 14.4\%		& 13.8\% 			& 28.2\%	\\ 
\textbf{Description} 	& 42.8\% 		& 21.0\%  			& 63.8\%	\\ 
\textbf{YesNo} 		& 2.9\% 		& 5.1\%   			& 8.0\%	\\
\hline
\textbf{Total} 		& 60.1\% 		& 39.9\%   			& 100.0\%	\\
\hline
\end{tabular}
\caption{The distribution of question types in DuReader is similar to (but different from) the Pilot Study (Table 3), largely because of duplicates. The duplicates were removed from DuReader (but not from the Pilot Study) to reduce the burden on the annotators.}
\label{table:query_dist}
\end{table}

\begin{table*}[!t]\footnotesize
\centering
\small
\begin{tabular}{lp{1.5\columnwidth}}
\hline
	\textbf{Question} & 学士服颜色 / What are the colors of academic dresses? \\
	\textbf{Question Type} & \emph{Entity-Fact}	\\
	\textbf{Answer 1} & \textbf{[绿色, 灰色, 黄色, 粉色]}：农学学士服绿色，理学学士服灰色，工学学士服黄色，管理学学士服灰色，法学学士服粉色，文学学士服粉色，经济学学士服灰色。 /\\
	{} & \textbf{[green, gray, yellow, pink]} Green for Bachelor of Agriculture, gray for Bachelor of Science, yellow for Bachelor of Engineering, gray for Bachelor of Management, pink for Bachelor of Law, pink for Bachelor of Art, gray for Bachelor of Economics\\
	\textbf{Document 1} & 农学学士服绿色，理学学士服灰色， ... ，确定为文、理、工、农、医、军事六大类，与此相应的饰边颜色为粉、灰、黄、绿、白、红六种颜色。	\\
	... & {} \\
	\textbf{Document 5} & 学士服是学士学位获得者在学位授予仪式上穿戴的表示学位的正式礼服， ... ，男女生都应着深色皮鞋。 \\
\hline
	\textbf{Question} & 智慧牙一定要拔吗 / Do I have to have my wisdom teeth removed \\
	\textbf{Question Type} & \emph{YesNo-Opinion}	\\
	\textbf{Answer 1} & \textbf{[Yes]}因为智齿很难清洁的原因，比一般的牙齿容易出现口腔问题，所以医生会建议拔掉 / \\
	{}				  & \textbf{[Yes]} The wisdom teeth are difficult to clean, and cause more dental problems than normal teeth do, so doctors usually suggest to remove them \\
	\textbf{Answer 2} & \textbf{[Depend]}智齿不一定非得拔掉，一般只拔出有症状表现的智齿，比如说经常引起发炎...  /\\
	{}				& \textbf{[Depend]} Not always, only the bad wisdom teeth need to be removed, for example, the one often causes inflammation ...\\
	\textbf{Document 1} & 为什么要拔智齿? 智齿好好的医生为什么要建议我拔掉?主要还是因为智齿很难清洁...	\\
	... & {} \\
	\textbf{Document 5} & 根据我多年的临床经验来说,智齿不一定非得拔掉.智齿阻生分好多种 ... \\
\hline
\end{tabular}
\caption{Examples from DuReader. Annotations for these questions include both the answers, as well as supporting sentences. } \label{table:qa_pair_samples}
\end{table*}

\subsubsection{Answer Annotation}
Crowd-sourcing was used to generate answers. Turkers were given a question and a set of relevant documents. He/she was then asked to write down answers in his/her own words by reading and summarizing the documents. If no answers can be found in the relevant documents, the annotator was asked to give an empty answer. If more than one answer can be found in the relevant documents, the annotator was asked to write them all down. In some cases, multiple answers were merged into a single answer, when it was determined that the multiple answers were very similar to one another.

Note that the answers to \emph{Entity} questions and \emph{YesNo} questions are more diverse. The answers to the \emph{Entity} questions include both the entities and the sentences containing them. See the first example in Table \ref{table:qa_pair_samples}. The bold words (i.e. green, gray, yellow, pink) are the entity answers to the question, and the sentences after the entities are the sentence containing them. The answers to the \emph{YesNo} questions include the opinion types (\emph{Yes}, \emph{No} or \emph{Depend}) as well as the supporting sentences. See the last example in Table \ref{table:qa_pair_samples}. The bold words (i.e. Yes and Depend) are the opinion types by following the supporting sentences. The second example shows that a simple yes-no question isn't so simple. The answer can be almost anything, including not only \emph{Yes} and \emph{No}, but also \emph{Depends}, depending on context (supporting sentences). 

\subsubsection{Quality Control}
Quality control is important because of the size of this project: $51,408$ man-hours distributed over about $800$ workers and $52$ experts. 


We have an internal crowdsourcing platform and annotation guidelines to annotate data. When annotating answers, workers are hired to create the answers and experts are hired to validate the answer quality. The workers will be hired if they pass an examine on a small dataset. The accuracy of workers' annotation should be higher than $95\%$ (judged by the experts). Basically, there are three rounds for answer annotations: (1) the workers will give the answers to the questions after reading the relevant documents. (2) the experts will review all answers created by the workers, and they will correct the answers if they consider that the answers are wrong. The accuracy (judged by the experts) of answers by the workers is around $90\%$. (3) The dataset is divided into $20$ groups according to the workers and experts who annotate the data. $5\%$ of data will be sampled from each group. The sampled data in each group will be further checked again by other experts. If the accuracy is lower than $95\%$, the corresponding workers and the experts need to revise the answers again. The loop will end until the overall accuracy reaches $95\%$. 


\subsubsection{Training, Development and Test Sets}
In order to maximize the reusability of the dataset, we provide a predefined split of the dataset into training, development and test sets. The training, development and test sets consist of $181K$, $10K$ and $10K$ questions, $855K$, $45K$ and $46K$ documents, $376K$, $20K$ and $21K$ answers, respectively.

\subsection{DuReader is (Relatively) Challenging}

Figures 1-2 illustrate some of the challenges of DuReader. 

\begin{figure}[t]
	\begin{center}
		\includegraphics[width=0.5\textwidth] {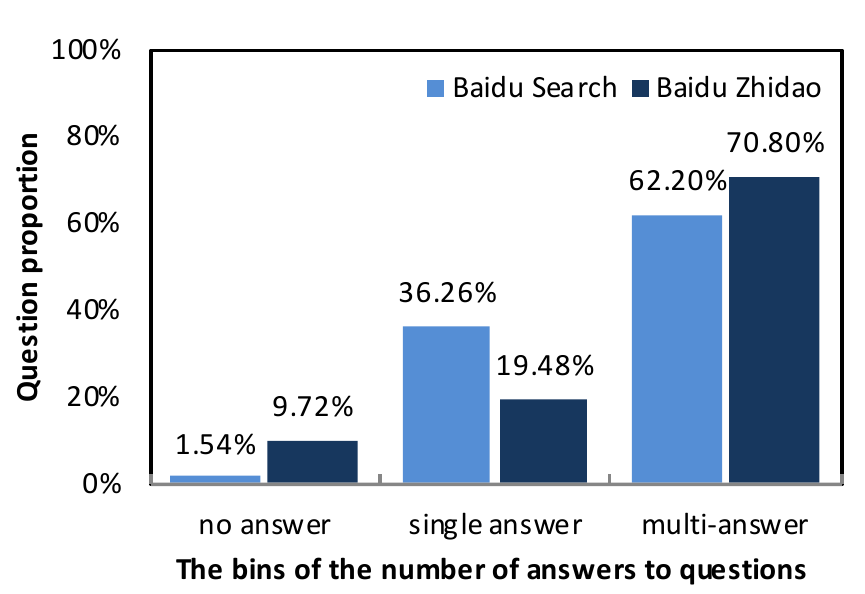} \\
		\caption{\label{figure:dr_answer_numbers} A few questions have one (and only one) answer,
        especially for Zhidao.}	
	\end{center}
\end{figure}

\textbf{The number of answers.} One might think that most questions would have one (and only one) answer, but Figure \ref{figure:dr_answer_numbers} shows that this is not the case, especially for Baidu Zhidao (70.8\% questions in Baidu Zhidao have multiple answers, while the number in Baidu Search is 62.2\%), where there is more room for opinions and subjectivity, and consequently, there is more room for diversity in the answer set. Meanwhile, we can see that 1.5\% of questions have zero answers in Baidu Search, but this number increases to 9.7\% in Baidu Zhidao. In the later case, no answer detection is a new challenge. 

\begin{figure}[t]
	\begin{center}
		\includegraphics[width=0.5\textwidth] {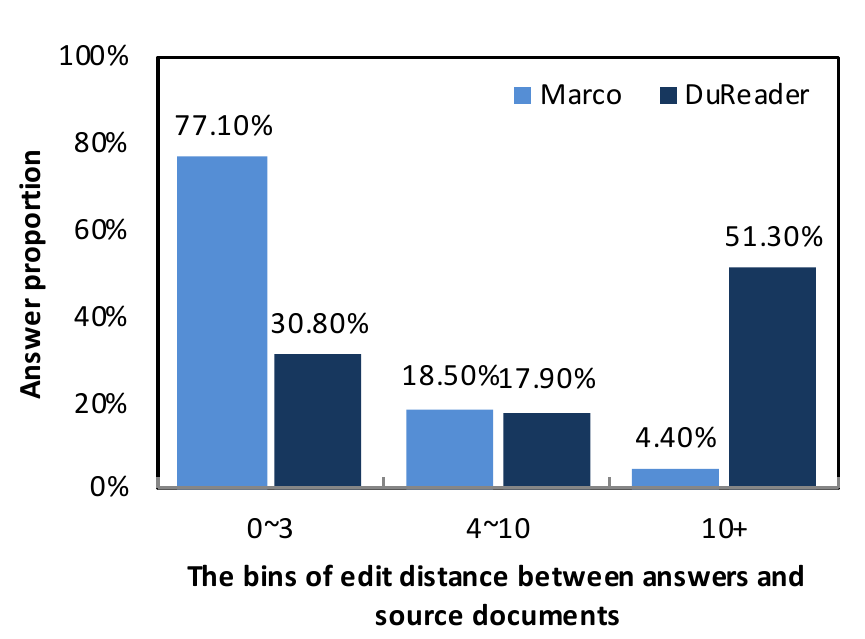} \\
		\caption{\label{figure:edit_distance} Span selection is unlikely to work well for DuReader because many of the answers are relatively far (in edit distance) from source documents (compared to MSMARCO). }
	\end{center}
\end{figure}

\textbf{The edit distance.} 
One might also have been tempted, based on prior work, to start with a span selection method, based on the success of such methods with previous datasets, many of which were designed for span selection, such as: SQuAD \cite{rajpurkar2016squad}, NewsQA \cite{trischler2017:newsqa} and TriviaQA \cite{joshi2017triviaqa}. However, this may not work well on DuReader, since the difference between the human generated answers and the source documents is large. To measure the difference, we use as an approximate measurement the minimum edit distance (MED) between the answers generated by human and the source documents\footnote{Here MED is the minimum edit distance between the answer and any consecutive span in the source document.}. A large MED means that an annotator needs to make more efforts on summarizing and paraphrasing the source documents to generate an answer, instead of just copying words from the source documents. Figure 2 compares DuReader and MS-MARCO in terms of MED, and suggests that span selection is unlikely to work well for DuReader where many of the answers are relatively far from source documents compared to MSMARCO. Note that the MED of SQuAD, NewsQA and TriviaQA should be zero.

\textbf{The document length.} In DuReader, questions tend to be short (4.8 words on average) compared to answers (69.6 words), and answers tend to be short compared to documents (396 words on average).  The documents in DuReader are 5x longer than documents in MS-MARCO~\cite{nguyen2016marco}. The difference is due to a design decision to provide unabridged documents (as opposed to paragraphs). We believe unabridged documents may be helpful because there may be useful clues throughout the document well beyond a single paragraph or a few passages.

\section{Experiments}

In this section, we implement and evaluate the baseline systems with two state-of-the-art models. Furthermore, with the rich annotations in our dataset, we conduct comprehensive evaluations from different perspectives. 

\begin{table*}[!h]
\centering
\begin{tabular}{c|cccccc}
\hline
\textbf{Systems}    & \multicolumn{2}{c}{\textbf{Baidu Search}} & \multicolumn{2}{c}{\textbf{Baidu Zhidao}} & \multicolumn{2}{c}{\textbf{All}} \\
                    		& BLEU-4\% & Rouge-L\% & BLEU-4\% & Rouge-L\% & BLEU-4\% & Rouge-L\% \\ \hline
\textbf{Selected Paragraph} & 15.8    & 22.6     & 16.5    & 38.3     & 16.4    & 30.2     \\ \hline
\textbf{Match-LSTM} 		& 23.1    & 31.2     & 42.5    & 48.0     & 31.9    & 39.2      \\
\textbf{BiDAF}      		& 23.1    & 31.1     & 42.2    & 47.5     & 31.8    & 39.0      \\ \hline\hline
\textbf{Human}        	 	& 55.1    & 54.4     & 57.1    & 60.7     & 56.1    & 57.4      \\ \hline
\end{tabular}
\caption{Performance of typical MRC systems on the DuReader.}
\label{table:perf_rc_sys}
\end{table*}

\begin{table}[]
\centering
\begin{tabular}{c|cc}
\hline
\textbf{}              	& BLEU-4\% & Rouge-L\% \\ \hline
\textbf{Gold Paragraph} & 31.7    & 61.3     \\ \hline
\textbf{Match-LSTM}	 	& 46.3    & 52.4     \\
\textbf{BiDAF}         	& 46.3    & 51.8     \\ \hline
\end{tabular}
\caption{Model performance with gold paragraph. The use of gold paragraphs could significantly boosts the overall performance.}
\label{table:performance_gold_para}
\end{table}

\begin{table*}[!h]
\centering
\begin{tabular}{c|cccccc}
\hline
\textbf{Question type}    & \multicolumn{2}{c}{\textbf{Description}} & \multicolumn{2}{c}{\textbf{Entity}} & \multicolumn{2}{c}{\textbf{YesNo}} \\
                    & BLEU-4\%         & Rouge-L\%       & BLEU-4\%         & Rouge-L\%       & BLEU-4\%         & Rouge-L\%          \\ \hline
\textbf{Match-LSTM} & 32.8           & 40.0          & 29.5           & 38.5          & 5.9            & 7.2             \\
\textbf{BiDAF}      & 32.6           & 39.7          & 29.8           & 38.4          & 5.5            & 7.5              \\ \hline\hline
\textbf{Human}      & 58.1           & 58.0          & 44.6           & 52.0          & 56.2           & 57.4             \\ \hline
\end{tabular}
\caption{Performance on various question types. Current MRC models achieve impressive improvements compared with the selected paragraph baseline. However, there is a large gap between these models and human.}
\label{table:perf_on_ans_type}
\end{table*}

\subsection{Baseline Systems} \label{section:baseline_system}

As we discussed in previous section, DuReader provides each question the full documents that contain multi-paragraphs or multi-passages, while previous work provides only a single paragraph~\cite{rajpurkar2016squad} or a few passages~\cite{nguyen2016marco} to extract or generate answers. The average length of each document is much longer than previous ones~\cite{nguyen2016marco}. If we directly apply the state-of-the-art MRC models that was designed for answer span selction, there will be efficiency issues. To improve both the efficiency of training and testing, our designed systems have two steps: (1) select one most related paragraph from each document, and (2) apply the state-of-the-art MRC models on the selected paragraphs. 

\subsubsection{Paragraph Selection}

In this paper, we apply simple strategies to select the most relevant paragraph from each document. In training stage, we select one paragraph from a document as the most relevant one, if the paragraph has the largest overlap with human generated answer. We select one most relevant paragraph for each document. Then, MRC models designed for answer span selection will be trained on these selected paragraphs. 

In testing stage, since we have no human generated answer, we select the most relevant paragraph that has the largest overlap with the corresponding question. Then, the trained MRC models designed for answer span selection will be applied on the these selected paragraphs. 

\subsubsection{Answer Span Selection}
We implement two typical state-of-the-art models designed for answer span selection as baselines. 

\textbf{Match-LSTM} Match-LSTM is a widely used MRC model and has been well explored in recent studies \cite{wang2017machine}. To find an answer in a paragraph, it goes through the paragraph sequentially and dynamically aggregates the matching of an attention-weighted question representation to each token of the paragraph. Finally, an answer pointer layer is used to find an answer span in the paragraph.

\textbf{BiDAF} BiDAF is a promising MRC model, and its improved version has achieved the best single model performance on SQuAD dataset \cite{seo2016bidirectional}. It uses both context-to-question attention and question-to-context attention in order to highlight the important parts in both question and context. After that, the so-called attention flow layer is used to fuse all useful information in order to get a vector representation for each position. 

\textbf{Implementation Details}
We randomly initialize the word embeddings with a dimension of $300$ and set the hidden vector size as $150$ for all layers. We use the Adam algorithm \cite{Diederik2014Adam} to train both MRC models with an initial learning rate of $0.001$ and a batch size of $32$.

\subsection{Results and Analysis}

\begin{table*}[]
\centering
\begin{tabular}{c|cccc}
\hline
\textbf{}					& \multicolumn{2}{c}{\textbf{Fact}} & \multicolumn{2}{c}{\textbf{Opinion}}	\\
\textbf{}					&	BLEU-4\%	& Rouge-L\% & BLEU-4\% & Rouge-L\% \\
\hline
\textbf{Opinion-unaware}  	&  6.3    	& 8.3	 & 5.0	 & 7.1     \\
\textbf{Opinion-aware}  	& 12.0    	& 13.9   & 8.0   & 8.9		\\ 
\hline
\end{tabular}
\caption{Performance of opinion-aware model on \emph{YesNo} questions.}
\label{table:perf_yn_opin}
\end{table*}

We evaluate the reading comprehension task via character-level BLEU-4 \cite{papineni2002bleu} and Rouge-L \cite{lin2004text}, which are widely used for evaluating the quality of language generation. The experimental results on test set are shown in Table \ref{table:perf_rc_sys}. For comparison, we also evaluate the Selected Paragraph that has the largest overlap with the question among all documents. We also assess human performance by involving a new annotator to annotate on the test data and treat his first answer as the prediction.

The results demonstrate that current reading comprehension models can achieve an impressive improvement compared with the selected paragraph baseline, which approves the effectiveness of these models. However, there is still a large performance gap between these models and human. An interesting discovery comes from the comparison between results on Baidu Search and Baidu Zhidao data. We find that the reading comprehension models get much higher score on Zhidao data. This shows that it is much harder for the models to comprehend open-domain web articles than to find answers in passages from a question answering community. In contrast, the performance of human beings on these two datasets shows little difference, which suggests that human’s reading skill is more stable on different types of documents.

As described in Section \ref{section:baseline_system}, the most relevant paragraph of each document is selected based on its overlap with the corresponding question during testing stage. To analyze the effect of paragraph selection and obtain an upper bound of the baseline MRC models, we re-evaluate our systems on the gold paragraphs, each of which is selected if it has the largest overlap with the human generated answers in a document. The experiment results have been shown in Table~\ref{table:performance_gold_para}. Comparing Table \ref{table:performance_gold_para} with Table \ref{table:perf_rc_sys}, we can see that the use of gold paragraphs could significantly boosts the overall performance. Moreover, directly using the gold paragraph can obtain a very high Rouge-L score. It meets the exception, because each gold paragraph is selected based on recall that is relevant to Rouge-L. Though, we find that the baseline models can get much better performance with respect to BLEU, which means the models have learned to select the answers. These results show that paragraph selection is a crucial problem to solve in real applications, while most current MRC datasets suppose to find the answer in a small paragraph or passage. In contrast, DuReader provides the full body text of each document to stimulate the research in a real-world setting. 

To gain more insight into the characteristics of our dataset, we report the performance across different question types in Table \ref{table:perf_on_ans_type}. We can see that both the models and human achieve relatively good performance on description questions, while \emph{YesNo} questions seem to be the hardest to model. We consider that description questions are usually answered with long text on the same topic. This is preferred by BLEU or Rouge. However, the answers to \emph{YesNo} questions are relatively short, which could be a simple \emph{Yes} or \emph{No} in some cases. 

\subsection{Opinion-aware Evaluation}
Considering the characteristics of \emph{YesNo} questions, we found that it’s not suitable to directly use BLEU or Rouge to evaluate the performance on these questions, because these metrics could not reflect the agreement between answers. For example, two contradictory answers like "You can do it" and "You can't do it" get high agreement scores with these metrics. A natural idea is to formulate this subtask as a classification problem. However, as described in Section \ref{section:dureader_dataset}, multiple different judgments could be made based on the evidence collected from different documents, especially when the question is of opinion type. In real-world settings, we don’t want a smart model to give an arbitrary answer for such questions as \emph{Yes} or \emph{No}.

To tackle this, we propose a novel opinion-aware evaluation method that requires the evaluated system to not only output an answer in natural language, but also give it an opinion label. We also have the annotators provide the opinion label for each answer they generated. In such cases, every answer is paired with an opinion label (\emph{Yes}, \emph{No} or \emph{Depend}) so that we can categorize the answers by their labels. Finally, the predicted answers are evaluated via Blue or Rouge against only the reference answers with the same opinion label. By using this opinion-aware evaluation method, a model that can predict a good answer in natural language and give it an opinion label correctly will get a higher score.

In order to classify the answers into different opinion polarities, we add a classifier. We slightly change the Match-LSTM model, in which the final pointer network layer is replaced with a fully connected layer. This classifier is trained with the gold answers and their corresponding opinion labels. We compare a reading comprehension system equipped with such an opinion classifier with a pure reading comprehension system without it, and the results are demonstrated in Table \ref{table:perf_yn_opin}. We can see that doing opinion classification does help under our evaluation method. Also, classifying the answers correctly is much harder for the questions of opinion type than for those of fact type.

\subsection{Discussion}
As shown in the experiments, the current state-of-the-art models still underperform human beings by a large margin on our dataset. There is considerable room for improvement on several directions.

First, there are some questions in our dataset that have not been extensively studied before, such as yes-no questions and opinion questions requiring multi-document MRC. New methods are needed for opinion recognition, cross-sentence reasoning, and multi-document summarization. Hopefully, DuReader’s rich annotations would be useful for study of these potential directions.

Second, our baseline systems employ a simple paragraph selection strategy, which results in great degradation of the system performance as compared to gold paragraph’s performance. It is necessary to design a more sophisticated paragraph ranking model for the real-world MRC problem.

Third, the state-of-the-art models formulate reading comprehension as a span selection task. However, as shown in previous section, human beings actually summarize answers with their own comprehension in DuReader. How to summarize or generate the answers deserves more research. 

Forth, as the first release of the dataset, it is far from perfection and it leaves much room for improvement. For example, we annotate only opinion tags for yes-no questions, we will also annotate opinion tags for description and entity questions. We would like to gather feedback from the community to improve DuReader continually.

Overall, it is necessary to propose new algorithms and models to tackle with real-world reading comprehension problems. We hope that the DuReader would be a good start for facilitating the MRC research.

\section{A Shared Task}
To encourage the exploration of more models from the research community, we organize an online competition\footnote{\url{https://ai.baidu.com/broad/leaderboard?dataset=dureader}}. Each participant can submit the result and evaluate the system performance at the online website. Since the release of the task, there are significant improvements over the baselines, For example, a team obtained 51.2 ROUGE-L on our dataset (when the paper was submitted). The gap between our BiDAF baseline model (with 39.0 ROUGE-L) and human performance (with 57.4 ROUGE-L) has been significantly reduced. It is expected that the remaining gap the system performances and human performance will be harder to close, but such efforts will lead to advances in machine reading comprehension. 

\section{Conclusion}
This paper announced the release of DuReader, a new dataset for researchers interested in machine reading comprehension (MRC). DuReader has three advantages over previous MRC datasets: (1) data sources (based on search logs and the question answering community), (2) question types (fact/ opinion \& entity/ description/ yes-no) and (3) scale (largest Chinese MRC dataset so far). 


We have made our dataset freely available and organize a shared competition to encourage the exploration of more models. Since the release of the task, we have already seen significant improvements from more sophisticated models.

\section*{Acknowledgements}
We would like to thank Dr. Kenneth Ward Church for his valuable suggestions and revisions on this paper, Prof. Sujian Li for her supports on this paper, and the anonymous reviewers for their helpful comments on this work.

\bibliography{acl}
\bibliographystyle{acl_natbib}

\end{CJK*}

\end{document}